\documentclass[lettersize,journal]{IEEEtran}

\usepackage{amsmath,amsfonts}
\usepackage{array}
\usepackage{multirow}
\usepackage{booktabs}
\usepackage{graphicx}
\usepackage[caption=false,font=normalsize,labelfont=sf,textfont=sf]{subfig}
\usepackage{textcomp}
\usepackage{stfloats}
\usepackage{url}
\usepackage{verbatim}
\usepackage{cite}
\usepackage{placeins}

\begin{document}

\title{Learning from Imperfect Teachers for Low-Resource Acoustic Generalization}

\author{Shuanglin Li,~\IEEEmembership{Member, IEEE},
        Ruxiao Qian,~\IEEEmembership{Member, IEEE},
        Jian Liu,~\IEEEmembership{Member, IEEE},
        Haijun Lin,~\IEEEmembership{Member, IEEE},
        Wenwu Wang,~\IEEEmembership{Fellow, IEEE},
        and Siyang Song$^*$,~\IEEEmembership{Member, IEEE}}

\maketitle

\begin{abstract}
Knowledge distillation (KD) improves low-resource acoustic learning by enriching
one-hot supervision with the softened predictive distribution of a fixed teacher
network. However, a teacher trained with limited or imbalanced annotations may
produce a biased distribution whose components are not uniformly reliable.
Although this distribution can still encode useful knowledge, direct full-distribution matching may also transfer teacher-induced biases, thereby distorting the student's decision boundary and degrading its generalization performance. To address this limitation, we propose Boundary-Anchored Mass-Partitioned Distillation (BA-MPD), a logit-based distillation objective composed of Boundary-Anchored Correction (BAC) and Mass-Partitioned Distillation (MPD). BAC addresses missing ground-truth labels in the set of the teacher’s top predictions by swapping the true label for the lowest-ranked entry of the set, thus keeping the mass and uncertainty of the set unchanged. MPD then distills this corrected distribution through separate losses that enforce relational consistency within the set, balance the mass between high- and low-confidence groups, and weight lower-confidence dependencies. Ultimately, BAC and MPD together suppress harmful ranking errors and noisy low-confidence details, while retaining all useful teacher information. Experiments on two acoustic benchmarks under multiple label budgets show that BA-MPD consistently improves over supervised-learning baselines and vanilla KD while remaining competitive with strong logit-based KD baselines. Cross-budget results further show that BA-MPD remains effective when the teacher and student models use mismatched label budgets, demonstrating its ability to exploit imperfect teachers across supervision gaps. Implementation available at: \url{https://github.com/ShuanglinLi/BA-MPD}.
\end{abstract}



\section{Introduction}
Low-resource acoustic learning remains challenging because reliable decision boundaries must be learned from limited labeled recordings across diverse acoustic environments and recording conditions, often with uneven class frequencies~\cite{heittola2020dcase,martin2021dcase,koutini2021receptive,
syed2024data}. In this regime, conventional one-hot labels provide only class-level supervision
by identifying the ground-truth class for each recording. These labels do not offer information about acoustic class relationships, uncertainty, or confusability among non-target classes~\cite{zhou2021rethinking,zhao2022decoupled}. This information proves especially valuable in low-resource acoustic settings, where few annotations make it hard to infer fine-grained class relationships from hard labels alone~\cite{chen2022htsat,chen2023beats}.

Knowledge distillation (KD)~\cite{hinton2015distilling} complements conventional one-hot label supervision by conveying the smoothed predictive distribution of a pre-trained, fixed teacher network to a student model. Unlike a one-hot label, which identifies only the ground-truth class, the teacher distribution assigns probabilities to all classes and can encode inter-class similarities,
plausible confusions among competing classes, and predictive uncertainty~\cite{gou2021knowledge,zhao2022decoupled}. Compared with hard labels, this distributional supervision provides a smoother and more informative training signal that can regularize boundaries of student decision in low-resource settings~\cite{hinton2015distilling,zhou2021rethinking}.

However, the reliability of such distributional supervision depends on the quality of the teacher from which it is derived. Under limited or imbalanced supervision, the teacher is trained on incomplete or skewed acoustic evidence, which can bias its softened outputs toward acoustically plausible but incorrect classes. Such biases can reflect unstable decision boundaries, spurious class correlations, or class-frequency bias~\cite{lukasik2021teachers,
stanton2021does,zhang2023blindly,cui2019class,zhang2025systematic,huang2025learn}. In this context, teacher soft predictions should not be regarded as uniformly trustworthy supervision. Since conventional full-distribution distillation aligns the entire teacher distribution with the student, it inevitably propagates both beneficial class relationships and undesirable elements, namely biased rankings and noisy probability estimates.

High-confidence errors from a teacher model are especially consequential when the ground-truth class is excluded from the prediction set, because direct distillation then encourages the student to reproduce a label-inconsistent ranking among the most likely classes. For a recording from an \emph{airport}, a high response for \emph{metro station} from a teacher can be informative, as both classes may share transportation-related and public-space acoustic patterns. The problem arises when \emph{street traffic} also receives high confidence while the ground-truth \emph{airport} class is excluded from the same prediction set. This ranking can be driven by generic traffic noise rather than airport-specific acoustic evidence, and direct distillation can transfer this spurious preference to the student model. However, the high-confidence predictions from a teacher model should not be discarded as a whole, since they may still encode useful acoustic relations. A core challenge, therefore, is to rectify label inconsistencies within the teacher's high-confidence prediction set, while simultaneously preserving the relational structure it encodes (\textbf{Challenge 1}).

Even after correcting label inconsistencies in the teacher's high-confidence set, the revised teacher distribution should not be treated as a single, monolithic target for distillation. The relative probabilities within the corrected high-confidence set reflect the teacher's ranking among plausible classes. Meanwhile, the proportion of total probability mass allocated to this set versus lower-confidence classes conveys the teacher's overall confidence in its corrected predictions. Although lower-confidence classes may retain some useful non-target information, their per-class probabilities are inherently noisier under limited supervision. Simply decomposing the full-distillation objective into separate terms does not fundamentally alter its behavior, as a result, relation learning remains modulated by the teacher's confidence distribution, and lower-confidence outputs are still matched on a class-by-class basis, preventing independent control over these signals. This motivates the second core challenge (i.e. \textbf{Challenge 2}): designing distinct objectives that separately govern relations within the high-confidence set, the mass split between high- and lower-confidence classes, and relations among lower-confidence classes.

To address these challenges, we propose Boundary-Anchored Mass-Partitioned
Distillation (BA-MPD), a structured distillation objective composed of Boundary-Anchored Correction (BAC) and Mass-Partitioned Distillation (MPD). BAC corrects a high-confidence prediction set when it excludes the ground-truth class through a boundary-level replacement. This retains the remaining high-confidence predictions, the total confidence assigned to the set, and the teacher's overall uncertainty. MPD then distills the corrected distribution through separate objectives capturing the relations within the high-confidence prediction set, the mass split between this set and lower-confidence classes, and weighted relations among lower-confidence classes. The main contributions of this work are summarized as follows:
\begin{itemize}
    \item We propose BA-MPD, a logit-based distillation objective for
    low-resource acoustic learning with imperfect teachers. BA-MPD treats
    teacher supervision as partially reliable and combines boundary-level
    correction with mass-partitioned distillation.

    \item We propose BAC, a conservative boundary-level correction that restores the ground-truth class to the teacher's high-confidence prediction set when it is absent. The correction retains the remaining high-confidence predictions, the total confidence assigned to the set, and the teacher's overall uncertainty.

    \item We introduce MPD, which uses separate objectives for relation
    matching within the corrected high-confidence prediction set, mass matching
    between this set and its complement, and weighted relation matching
    within the complement.

    \item We evaluate BA-MPD on two benchmarks under multiple label
    budgets. BA-MPD improves over supervised-learning baselines
    and vanilla KD while remaining competitive with strong logit-based KD
    baselines. Cross-budget, ablation, and teacher-reliability analyses further
    show its behavior under supervision gaps and the roles of BAC and MPD.
    Code, fixed splits, checkpoints, and teacher logits are available on this page
    \footnote{https://github.com/ShuanglinLi/BA-MPD}.
\end{itemize}

\section{Related Work}

\subsection{Knowledge Distillation}

Knowledge distillation transfers predictive knowledge from a teacher to a student by matching softened output distributions~\cite{hinton2015distilling}. Compared with one-hot labels, teacher soft targets can encode inter-class similarity, predictive uncertainty, and non-target dark knowledge~\cite{gou2021knowledge}. Conventional logit-based KD typically matches the full teacher distribution with a single Kullback–Leibler (KL) divergence based objective, implicitly treating all probability components as equally reliable.

Recent studies have addressed this limitation by decomposing, reweighting, or selectively using teacher supervision. Decoupled KD separates the target and non-target contributions~\cite{zhao2022decoupled}, while reliability-aware methods examine teacher correctness and the risk of blindly imitating unreliable teacher outputs~\cite{lukasik2021teachers,stanton2021does,zhang2023blindly,mishra2026correctness}. These efforts show the importance of structured teacher supervision, but they mainly operate at the sample, target, or full-distribution level, without explicitly addressing cases where the ground-truth label falls outside the teacher's top-ranked high-confidence predictions. Our work follows this structured view in the same-budget low-resource setting: BA-MPD corrects the teacher's top-\(k\) high-confidence predictions when they exclude the ground-truth label, rather than discarding such samples, and separates probability-mass matching between the corrected top-\(k\) set and the remaining classes from relation matching within each part.

\subsection{Low-Resource Acoustic Representation Learning}

Low-resource acoustic representation learning aims to build robust acoustic models from scarce labeled recordings under intra-class variability, background noise, recording-condition shifts, and class imbalance~\cite{heittola2020dcase,martin2021dcase,koutini2021receptive,syed2024data}. These challenges appear in data-efficient acoustic scene classification benchmarks such as DCASE Task 1~\cite{heittola2020dcase,martin2021dcase,kim2022rfn}, as well as in bioacoustic benchmarks with fine-grained and long-tailed categories ~\cite{hagiwara2023beans}.

Existing approaches improve low-resource acoustic representation learning through data
augmentation, pretraining, architecture design, domain-robust training, and
regularization. In contrast, we study how to use teacher soft predictions as
structured supervision when the teacher itself is trained under the same
low-resource label budget. In this setting, recording-condition shifts and
long-tailed class distributions can bias the teacher's high-confidence predictions toward acoustically similar or frequent classes. BA-MPD addresses
this problem at the distillation-objective level, without requiring additional annotations, class reweighting losses, or task-specific resampling strategies.

\begin{figure*}[t]
\centering
\includegraphics[width=\textwidth,
        height=0.30\textheight,
        keepaspectratio]{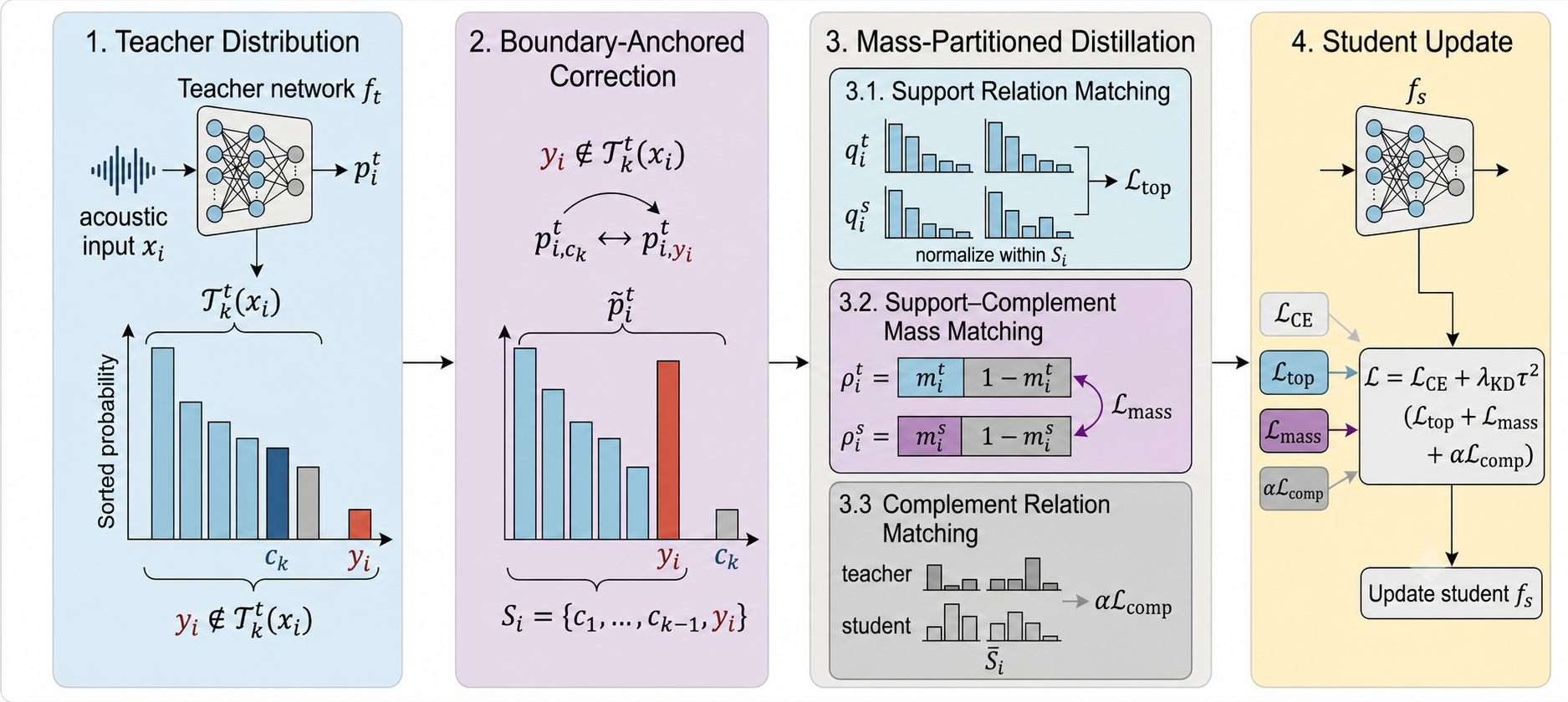}
\caption{Overview of Boundary-Anchored Mass-Partitioned Distillation (BA-MPD). The figure illustrates how BAC converts a teacher top-\(k\) support with label-support mismatch into a label-consistent anchored support \(S_i\) through a boundary-level label exchange, while MPD distills the anchored teacher distribution by separating support relation matching, support--complement mass matching, and complement relation matching. The resulting BA-MPD loss is combined with cross-entropy (CE) to update the student.}
\label{fig:ba_mpd}
\end{figure*}

\section{Preliminaries: Revisiting Knowledge Distillation}
\label{sec:preliminaries}

Knowledge distillation trains a student network by matching its
temperature-softened predictive distribution to that of a fixed teacher
network. Compared with one-hot labels that specify only the ground-truth class,
teacher soft predictions provide additional information about inter-class
similarity and predictive uncertainty. Given a mini-batch of labeled acoustic
samples \(\mathcal{B}=\{(x_i,y_i)\}_{i=1}^{B}\) from a \(C\)-class
classification task, \(x_i\) denotes an input recording and
\(y_i\in\{1,\ldots,C\}\) denotes its ground-truth class index. Let
\(z_i^t,z_i^s\in\mathbb{R}^{C}\) be the teacher and student pre-softmax logits
for \(x_i\), respectively. With temperature \(\tau\), the softened
class-probability distributions \(p_i^t,p_i^s\in\mathbb{R}^{C}\) are defined as:
\begin{equation}
p_i^t=\operatorname{softmax}(z_i^t/\tau),
\qquad
p_i^s=\operatorname{softmax}(z_i^s/\tau).
\end{equation}
where the elements \(p_{i,c}^t\) and \(p_{i,c}^s\) denote the teacher and student
probabilities assigned to class \(c\), and \(\tau\) controls the smoothness of the two distributions.

Vanilla KD minimizes the KL divergence between the teacher and student
distributions over all \(C\) classes:
\begin{equation}
\label{eq:vanilla_kd}
\mathcal{L}_{\mathrm{KD}}^{\mathrm{vanilla}}
=
\frac{\tau^2}{B}
\sum_{i=1}^{B}
\sum_{c=1}^{C}
p_{i,c}^t
\log
\frac{p_{i,c}^t}{p_{i,c}^s}.
\end{equation}
Here, the loss is averaged over the mini-batch, and \(\tau^2\) compensates for the
gradient-scale change induced by temperature softening.

Equation~\eqref{eq:vanilla_kd} matches all class probabilities within a single
full-space objective. To distinguish the teacher's high-confidence predictions
from its remaining responses, we first obtain the ordered indices of the
\(k\) classes with the largest teacher probabilities:
\begin{equation}
(c_1,\ldots,c_k)=\operatorname{TopK}(p_i^t,k),
\end{equation}
where
\(p_{i,c_1}^t\ge\cdots\ge p_{i,c_k}^t\). The teacher top-\(k\)
high-confidence support is then defined as:
\begin{equation}
\label{eq:support}
\mathcal{T}_k^t(x_i)=\{c_1,\ldots,c_k\}.
\end{equation}
Thus, \(c_k\) is the lowest-probability class within the support set of the teacher. The
complementary set contains the remaining \(C-k\) classes, and the two sets partition
the full class space. In the following sections, ``support'' refers to this set of top-\(k\) high-confidence classes unless otherwise specified.

A label-support mismatch occurs when
\(y_i\notin\mathcal{T}_k^t(x_i)\), meaning that the teacher's \(k\)
highest-probability classes exclude the ground-truth class. In this case,
vanilla KD still encourages the student to reproduce the teacher's
high-confidence ranking, even though that ranking is label-inconsistent. More
generally, its single full-space objective provides no independent control over
the relative probabilities within the support, the probability mass allocated
between the support and its complementary set, or the relative probabilities within
the complementary set. This motivates a structured strategy that first repairs the
label-support mismatch and then distills these three signals with separate
objectives.

\section{Methodology}
\label{sec:methodology}

This section presents Boundary-Anchored Mass-Partitioned Distillation (BA-MPD), a structured distillation objective for low-resource representation learning with imperfect teachers. We consider a fixed teacher trained under the same limited label budget as the student, rather than a fully reliable external oracle. As discussed in Section~\ref{sec:preliminaries}, such a teacher can provide useful class relations, but its support may exclude the ground-truth class, while
full-distribution KD ties support responses, complement responses, and their allocated mass into one objective. BA-MPD addresses these two issues by repairing label-support mismatch through BAC and distilling the corrected teacher distribution with separated MPD objectives.

\textbf{Notation.}  We follow the notation in Section~\ref{sec:preliminaries}. For each sample
\(x_i\), \(p_i^t\) and \(p_i^s\) denote the temperature-softened class-confidence distributions predicted by the fixed teacher and the student, respectively. The teacher top-\(k\) support is denoted by \(\mathcal{T}_k^t(x_i)\), with its ordered classes written as \((c_1,\ldots,c_k)\). The boundary class \(c_k\) is the lowest-probability class within this support. BAC produces an anchored teacher distribution \(\tilde{p}_i^t\) and an anchored support \(S_i\). MPD then partitions the class space into \(S_i\) and its complement \(\bar{S}_i=\{1,\ldots,C\}\setminus S_i\), where \(\bar{S}_i\) contains all classes outside the anchored support. \\

\textbf{Overview.} As illustrated in Figure~\ref{fig:ba_mpd}, BA-MPD consists of two components,
Boundary-Anchored Correction (BAC) and Mass-Partitioned Distillation (MPD). For each training sample, BAC examines the teacher distribution \(p_i^t\) through its top-\(k\) support \(\mathcal{T}_k^t(x_i)\). If \(y_i\) is absent from this support, BAC replaces the boundary class \(c_k\) with \(y_i\) and constructs the anchored teacher distribution by reassigning the probabilities associated with these two classes; otherwise, both the teacher distribution and its support remain unchanged. The output of BAC is an anchored teacher distribution \(\tilde{p}_i^t\) and an anchored support \(S_i\). MPD then uses \(\tilde{p}_i^t\), the student distribution \(p_i^s\), \(S_i\), and its complement \(\bar{S}_i\) to construct support relation matching \(\mathcal{L}_{\mathrm{top}}\), support--complement mass matching \(\mathcal{L}_{\mathrm{mass}}\), and complement relation matching
\(\mathcal{L}_{\mathrm{comp}}\). These three terms form the BA-MPD distillation loss and are combined with supervised cross-entropy to train the student. The following subsections describe BAC, MPD, and the overall training objective in detail.

\subsection{Boundary-Anchored Correction}
\label{subsec:boundary_anchor}

BAC addresses label-support mismatch in the teacher top-\(k\) support, where
the ground-truth class \(y_i\) is absent from the teacher's high-confidence
predictions. Instead of correcting all teacher errors or reshaping the full
distribution, BAC applies a boundary-level repair that makes the local support
label-consistent while preserving the leading teacher responses.

We use the ordered teacher support
\(\mathcal{T}_k^t(x_i)=\{c_1,\ldots,c_k\}\) defined in
Section~\ref{sec:preliminaries}, where
\(p_{i,c_1}^t\ge\cdots\ge p_{i,c_k}^t\). The last element \(c_k\) is treated as
the boundary class, i.e., the lowest-probability class within the teacher
top-\(k\) support. BAC outputs the anchored support \(S_i\) as:
\begin{equation}
\label{eq:anchored_topk}
S_i =
\begin{cases}
\mathcal{T}_k^t(x_i),
& y_i\in\mathcal{T}_k^t(x_i),\\
\{c_1,\ldots,c_{k-1},y_i\},
& y_i\notin\mathcal{T}_k^t(x_i).
\end{cases}
\end{equation}
When \(y_i\in\mathcal{T}_k^t(x_i)\), the teacher support is already
label-consistent and BAC applies no correction by setting
\(\tilde{p}_i^t=p_i^t\). When \(y_i\notin\mathcal{T}_k^t(x_i)\), the
ground-truth class lies outside the teacher support, and BAC replaces only the
boundary class \(c_k\) with \(y_i\) in the anchored support.

This support-level replacement is implemented by assigning the boundary
probability to \(y_i\) and assigning the original probability of \(y_i\) to
\(c_k\):
\begin{equation}
\label{eq:boundary_anchor}
\tilde{p}_{i,c}^{t} =
\begin{cases}
p_{i,c_k}^{t}, & c=y_i,\\
p_{i,y_i}^{t}, & c=c_k,\\
p_{i,c}^{t}, & c\notin\{y_i,c_k\}.
\end{cases}
\end{equation}
Thus, \(S_i\) always contains \(k\) classes. When correction is applied, its
total probability mass equals that of the original teacher top-\(k\) support,
while the leading classes \(c_1,\ldots,c_{k-1}\) and their probabilities remain
unchanged. This gives a conservative support-level repair rather than a
full-distribution rectification. The next paragraph explains why this repair is
anchored at the boundary class \(c_k\).

\textbf{Boundary repair rationale.} The boundary choice follows from the ordered teacher support \(\mathcal{T}_k^t(x_i)=\{c_1,\ldots,c_k\}\), where
\(p_{i,c_1}^t\ge\cdots\ge p_{i,c_k}^t\). Since \(c_k\) has the smallest
probability among the teacher's top-\(k\) classes, \(p_{i,c_k}^t\) is the minimum probability assigned to a class that is still included in the teacher's support. When \(y_i\notin\mathcal{T}_k^t(x_i)\), replacing \(c_k\) with \(y_i\) therefore places the ground-truth class at the boundary of the anchored support, while keeping the higher-probability classes \(c_1,\ldots,c_{k-1}\) unchanged.

To see why the boundary class is the conservative choice, consider replacing a
generic teacher support class \(c_j\in\mathcal{T}_k^t(x_i)\) with \(y_i\), while keeping the operation restricted to the two class coordinates \(y_i\) and
\(c_j\). Let \(\hat{p}_i^{t,(j)}\) denote the resulting distribution. The total
absolute change caused by this two-class reassignment is:
\begin{equation}
\begin{aligned}
D_j
&=
\sum_{c=1}^{C}
\left|
\hat{p}_{i,c}^{t,(j)}-p_{i,c}^{t}
\right|  \\
&=
\left|p_{i,c_j}^{t}-p_{i,y_i}^{t}\right|
+
\left|p_{i,y_i}^{t}-p_{i,c_j}^{t}\right|  \\
&=
2\left(p_{i,c_j}^{t}-p_{i,y_i}^{t}\right),
\end{aligned}
\end{equation}
where the last equality holds because \(p_{i,c_j}^{t}\ge p_{i,c_k}^{t}\ge p_{i,y_i}^{t}\) when \(y_i\notin\mathcal{T}_k^t(x_i)\). Since \(c_k\) has the smallest probability among all \(c_j\in\mathcal{T}_k^t(x_i)\), choosing \(c_k\) gives the smallest \(D_j\) within this two-class repair family.

This repair also preserves the teacher's probability values. BAC only reassigns
two existing probabilities, so the entropy of the teacher distribution remains
unchanged. Moreover, under the corrected distribution \(\tilde{p}_i^t\), the probability mass assigned to the anchored support equals the original teacher top-\(k\) mass:
\begin{equation}
\sum_{c\in S_i}\tilde{p}_{i,c}^t
=
\sum_{c\in\mathcal{T}_k^t(x_i)}p_{i,c}^t .
\end{equation}
Therefore, BAC repairs the label-support mismatch without correcting all
teacher errors or reshaping the full distribution. It makes the top-\(k\)
support label-consistent while preserving the teacher's uncertainty, the
original top-\(k\) mass, and the leading teacher classes.

However, BAC only specifies the label-consistent anchored support used after correction; it does not determine how the anchored distribution \(\tilde{p}_i^t\) should supervise the student. We therefore introduce MPD to distill the anchored teacher signal with separate objectives for the support region and its complement.

\subsection{Mass-Partitioned Distillation}
\label{subsec:mpd}

After boundary anchoring, each input recording \(x_i\) is associated with an
anchored teacher distribution \(\tilde{p}_i^t\) and a label-consistent anchored
support \(S_i\). MPD distills this anchored teacher signal through the
support--complement partition \((S_i,\bar{S}_i)\), where
\(\bar{S}_i=\{1,\ldots,C\}\setminus S_i\) contains all classes outside the
anchored support. Rather than matching \(\tilde{p}_i^t\) as a single
\(C\)-class target, MPD separates three factors: normalized class relations
within \(S_i\), probability mass allocation between \(S_i\) and \(\bar{S}_i\),
and normalized class relations within \(\bar{S}_i\). An exact decomposition of full-distribution KL would still weight the normalized relation terms by the teacher mass assigned to the support and complement. MPD instead matches this mass separately and removes the corresponding mass weighting from relation matching, allowing support relations, support--complement mass, and complement relations to be controlled independently. The complement relation term is further controlled by \(\alpha\) because its class-wise responses can be less stable under limited supervision.
 
\noindent
\textbf{Support relation matching.}
The first MPD objective focuses on the relative class relations within the
anchored support \(S_i\). After BAC, \(S_i\) contains the ground-truth class while retaining the teacher's leading responses. We normalize the teacher and student probabilities within \(S_i\) to isolate their relative probability structure from the total mass assigned to the support:
\begin{equation}
q_{i,c}^{t}=
\frac{\tilde{p}_{i,c}^{t}}
{\sum_{r\in S_i}\tilde{p}_{i,r}^{t}},
\qquad
q_{i,c}^{s}=
\frac{p_{i,c}^{s}}
{\sum_{r\in S_i}p_{i,r}^{s}},
\quad c\in S_i .
\end{equation}
where \(q_i^t\) and \(q_i^s\) are normalized distributions over the
\(k\)-class support \(S_i\). The support relation matching loss is defined as:
\begin{equation}
\mathcal{L}_{\mathrm{top}}
=
\frac{1}{B}\sum_{i=1}^{B}
\sum_{c\in S_i}
q_{i,c}^{t}
\log
\frac{q_{i,c}^{t}}{q_{i,c}^{s}} .
\end{equation}
This objective encourages the student to match the teacher's normalized
relative probability structure within the anchored support. Since \(q_i^t\) and \(q_i^s\) are normalized inside \(S_i\), \(\mathcal{L}_{\mathrm{top}}\) does not determine how much total probability mass should be assigned to the support. This mass allocation is handled by the next objective.

\noindent
\textbf{Support--complement mass matching.}
Beyond the normalized relations within \(S_i\), the anchored teacher
distribution also specifies how much probability mass is assigned to the
support compared with the remaining classes. MPD captures this coarse
confidence allocation by aggregating the teacher and student probabilities into
two-dimensional mass distributions:
\begin{equation}
m_i^t=\sum_{c\in S_i}\tilde{p}_{i,c}^t,
\qquad
m_i^s=\sum_{c\in S_i}p_{i,c}^s ,
\end{equation}
\begin{equation}
\rho_i^t=\left[m_i^t,\,1-m_i^t\right],
\qquad
\rho_i^s=\left[m_i^s,\,1-m_i^s\right].
\end{equation}
where \(\rho_i^t,\rho_i^s\in\mathbb{R}^{2}\) describe the mass assigned to two
parts of the class space: the first component corresponds to \(S_i\), and the
second component corresponds to \(\bar{S}_i\). The mass matching loss is:
\begin{equation}
\mathcal{L}_{\mathrm{mass}}
=
\frac{1}{B}\sum_{i=1}^{B}
\sum_{r\in\{1,2\}}
\rho_{i,r}^{t}
\log
\frac{\rho_{i,r}^{t}}{\rho_{i,r}^{s}} .
\end{equation}
This objective aligns the student's support--complement mass allocation with
the anchored teacher distribution. After matching the support relations and the
support--complement mass, the remaining teacher signal lies in the relative
pattern among the complement classes.

\noindent
\textbf{Complement relation matching.}
The complement region \(\bar{S}_i\) contains classes outside the anchored
support. Although these classes receive lower teacher probabilities, their
relative pattern may still contain complementary dark knowledge about
non-dominant acoustic classes. To transfer this information without mixing it
with the support mass, we normalize the teacher and student distributions
within \(\bar{S}_i\):
\begin{equation}
\bar{q}_{i,c}^{t}=
\frac{\tilde{p}_{i,c}^{t}}
{\sum_{r\in \bar{S}_i}\tilde{p}_{i,r}^{t}},
\qquad
\bar{q}_{i,c}^{s}=
\frac{p_{i,c}^{s}}
{\sum_{r\in \bar{S}_i}p_{i,r}^{s}},
\quad c\in \bar{S}_i .
\end{equation}
where \(\bar{q}_i^t\) and \(\bar{q}_i^s\) are normalized distributions over the
\((C-k)\)-class complement \(\bar{S}_i\). The complement relation matching loss
is defined as:
\begin{equation}
\mathcal{L}_{\mathrm{comp}}
=
\frac{1}{B}\sum_{i=1}^{B}
\sum_{c\in \bar{S}_i}
\bar{q}_{i,c}^{t}
\log
\frac{\bar{q}_{i,c}^{t}}{\bar{q}_{i,c}^{s}} .
\end{equation}
Because complement responses are lower-confidence and their class-wise values
can be less stable under low-resource supervision, we treat this term as
auxiliary and control its contribution with a weighting coefficient \(\alpha\).
Together, \(\mathcal{L}_{\mathrm{top}}\), \(\mathcal{L}_{\mathrm{mass}}\), and
\(\mathcal{L}_{\mathrm{comp}}\) form the complete MPD objective.

\noindent
\subsection{BA-MPD Training Objective}
\label{subsec:objective}

Combining the three region-wise objectives, the MPD loss for the anchored
teacher signal is defined as:
\begin{equation}
\mathcal{L}_{\mathrm{MPD}}
=
\mathcal{L}_{\mathrm{top}}
+
\mathcal{L}_{\mathrm{mass}}
+
\alpha \mathcal{L}_{\mathrm{comp}} 
\end{equation}
Here, \(\alpha\) controls the contribution of complement relation matching. The
student is trained with supervised cross-entropy and BA-MPD distillation:
\begin{equation}
\mathcal{L}
=
\mathcal{L}_{\mathrm{CE}}
+
\lambda_{\mathrm{KD}}\tau^2
\mathcal{L}_{\mathrm{MPD}} 
\end{equation}
Here, \(\mathcal{L}_{\mathrm{CE}}\) is the cross-entropy loss with the ground-truth label, \(\lambda_{\mathrm{KD}}\) balances supervised and distillation losses, and \(\tau^2\) follows the standard temperature-scaled KD formulation.

\section{Experiments}

This section evaluates BA-MPD on acoustic scene classification and
bioacoustic recognition under multiple label budgets. We first describe the
datasets, low-resource protocols, implementation details, evaluation metrics,
and comparison methods. We then present same-budget comparisons, ablation and
sensitivity studies, cross-budget transfer results, and analyses of teacher
reliability to assess the overall effectiveness of BA-MPD and examine the
contributions of its main design components.

\subsection{Datasets}

We evaluate the proposed BA-MPD on two complementary acoustic recognition benchmark datasets:
\begin{itemize}
\item \textbf{TAU Urban Acoustic Scenes}~\cite{heittola2020dcase} is a standard acoustic scene classification benchmark used in the DCASE Task 1 series. It contains recordings from multiple urban acoustic environments and is commonly used to evaluate robust acoustic scene recognition under shifts in devices and recording conditions.

\item \textbf{BEANS-CBI}~\cite{hagiwara2023beans} is a subset of the BEANS benchmark that contains bird vocalization recordings from 264 categories and naturally follows a long-tailed class distribution. It offers a complementary bioacoustic scenario where class imbalance and fine-grained acoustic confusability make low-resource recognition particularly challenging.
\end{itemize}

\textbf{Low-resource protocol.} For TAU Urban Acoustic Scenes, we follow the official low-resource setting of DCASE 2024 Task 1~\cite{schmid2024dcase}. For BEANS-CBI, since no predefined low-resource splits are provided, we create nested class-stratified subsets from the official training split with label budgets of 10\%, 25\%, and 50\%. For each class, we sample a proportion of the budget for training examples while retaining at least one example. The official validation and test splits remain unchanged.

\subsection{Experimental Setup and Implementation Details}

\noindent \textbf{Teacher and student networks.} For TAU Urban Acoustic Scenes, the student follows the official DCASE 2024 Task 1 CP-Mobile baseline architecture~\cite{schmid2024dcase}. The teacher is a PaSST-based acoustic model~\cite{koutini2021passt} trained on the same low-resource split, following the standard acoustic-scene distillation setup in~\cite{schmid2023distilling}, rather than a full-data privileged teacher. For BEANS-CBI, both teacher and student use the PANNs CNN14 architecture~\cite{kong2020panns}. For each budget and split seed, the CNN14 teacher is trained from scratch on the corresponding low-resource training subset and then kept fixed for student training. In both benchmarks, the teacher is fixed after low-resource task training, and its logits are exported once and shared by all KD-based methods under the same benchmark, budget, and split seed. Therefore, performance differences among KD methods mainly reflect the distillation objective rather than teacher selection, student architecture, or logit recomputation.

\noindent
\textbf{Training objective and hyper-parameter settings.}
The CE baseline trains the student with cross-entropy only. KD-based methods train the student using cross-entropy plus the corresponding distillation objective, while sharing the same teacher logits and training protocol across benchmarks and budgets. Supervised regularizers are not combined with KD in the main comparison and are evaluated as non-distillation baselines. For KD-based methods, we set the KD loss weight to \(\lambda_{\mathrm{KD}}=0.3\) and use \(\tau=2.0\) unless the baseline specifies a curriculum temperature schedule. BA-MPD introduces two method-specific hyperparameters: the support size \(k\) and the complement-region relation weight \(\alpha\). We set \(\alpha=2.0\), use \(k=2\) for TAU Urban Acoustic Scenes and \(k=5\) for BEANS-CBI as default value, and study the sensitivity to \(k\) in the ablation analysis.

\noindent \textbf{Other implementation details.} For TAU Urban Acoustic Scenes, we follow the official DCASE 2024 Task 1 CP-Mobile training recipe~\cite{schmid2024dcase}. Student models are trained for 150 epochs using AdamW~\cite{loshchilov2019decoupled} with a batch size of 256, a peak learning rate of \(5\times10^{-3}\), and a weight decay of \(10^{-4}\). The learning rate is linearly warmed up over the first 2,000 optimization steps and then decayed using a cosine schedule. All KD-based methods use a common 10-epoch supervised warm-up, followed by a 10-epoch linear increase of the distillation weight to its specified value.

For BEANS-CBI, audio is resampled to 32~kHz and represented using 64-bin log-mel spectrograms. The CNN14 models are trained for 20 epochs using AdamW with a batch size of 32, an initial learning rate of \(3\times10^{-4}\), and a weight decay of \(10^{-4}\). The learning rate is linearly warmed up over 200 optimization steps and then follows cosine decay, while the distillation loss is applied throughout training. Within each benchmark and label budget, all
comparison methods use the same preprocessing pipeline, optimization settings, model-selection criterion, and set of random seeds. Checkpoints are selected on the validation split using the primary metric of the corresponding benchmark, and the reported results are averaged over three independent runs. Each training run is conducted on a single NVIDIA GeForce RTX 4090 GPU. BAC is deterministic and introduces no learnable parameters or additional training schedule.

\noindent \textbf{Evaluation metrics.} For TAU Urban Acoustic Scenes, we follow the official DCASE Task 1 protocol and use macro-average scene classification accuracy as the primary metric, with seen- and unseen-device accuracies reported for device-shift analysis. For BEANS-CBI, we use macro accuracy, equivalent to balanced accuracy, as the primary metric due to its 264-class long-tailed label space, and additionally report macro-F1 and micro accuracy.

\subsection{Comparison Methods}

We organize the main controlled comparisons into two groups: logit-based KD
methods and supervised low-resource regularization baselines. We additionally
discuss feature-based KD, self-supervised and unsupervised representation learning methods as orthogonal low-resource strategies. \\

\noindent
\textbf{Logit-based KD methods.}
We compare BA-MPD with the following representative logit-based KD baselines. Vanilla KD~\cite{hinton2015distilling} minimizes the full-space KL divergence between the softened teacher and student output distributions. CTKD~\cite{li2023curriculum} adjusts the distillation temperature through a curriculum schedule. DKD~\cite{zhao2022decoupled} separates target-class and non-target-class distillation, while BTKD~\cite{zhang2024can} accounts for potentially biased teacher predictions during knowledge transfer. Since BA-MPD is designed to operate on teacher predictive distributions, we restrict the main comparison to logit-based KD methods. Feature-based KD methods, such as FitNets~\cite{romero2015fitnets} and RKD~\cite{park2019relational}, transfer intermediate representations and may introduce architecture-dependent interfaces. We therefore exclude them from the controlled comparison, which focuses on how the same fixed teacher predictions are distilled at the output level. \\

\noindent \textbf{Supervised low-resource regularization baselines.}
We also include supervised baselines that do not use teacher predictions. In addition to the cross-entropy baseline, we include common supervised regularization controls: label smoothing~\cite{szegedy2016rethinking}, mixup~\cite{zhang2018mixup}, and their combination for TAU Urban Acoustic Scenes, and label smoothing, focal loss~\cite{lin2017focal}, and their combination for the long-tailed BEANS-CBI benchmark. These baselines help verify whether the gains of BA-MPD arise from structured distillation of teacher distributions rather than from supervised regularization alone. Self-supervised and pretrained audio representations, such as wav2vec 2.0~\cite{baevski2020wav2vec}, HuBERT~\cite{hsu2021hubert}, and BEATs~\cite{chen2023beats}, are important low-resource acoustic learning strategies. However, they primarily change the representation or initialization available to the model, whereas our goal is to study how a fixed imperfect teacher distribution should be distilled under a fixed labeled target budget. We therefore treat representation pretraining as an orthogonal direction rather than a direct logit-distillation baseline.

\subsection{Main Results}

In this section, we evaluate BA-MPD under the same-budget low-resource setting. All methods are compared using the same teacher, student architecture, data split, and optimization protocol within each benchmark and label budget. We include both supervised regularization baselines and representative logit-based KD methods. The analysis focuses on the primary metric of each benchmark, while additional metrics are used to examine device-shift robustness on TAU Urban Acoustic Scenes and class-imbalance behavior on BEANS-CBI.

\begin{table*}[t]
\centering
\caption{Main results on TAU Urban Acoustic Scenes under different low-resource budgets. Reference Teacher denotes the split-specific PaSST teacher evaluated on the official test split and kept fixed for all KD methods within each budget. Student results are averaged over three random seeds and reported as macro-average scene classification accuracy, seen-device accuracy, and unseen-device accuracy (\%). $\Delta$ denotes the absolute improvement of BA-MPD over vanilla KD.}
\label{tab:tau_main_results}
\footnotesize
\setlength{\tabcolsep}{3.2pt}
\renewcommand{\arraystretch}{1.08}
\begin{tabular}{ccccccccccc}
\toprule
\multirow{3}{*}{Manner} &
\multirow{3}{*}{Method} &
\multicolumn{9}{c}{TAU Urban Acoustic Scenes: PaSST teacher $\rightarrow$ CP-Mobile student} \\
\cmidrule(lr){3-11}
& & \multicolumn{3}{c}{5\%} & \multicolumn{3}{c}{10\%} & \multicolumn{3}{c}{25\%} \\
\cmidrule(lr){3-5} \cmidrule(lr){6-8} \cmidrule(lr){9-11}
& & Macro & Seen & Unseen & Macro & Seen & Unseen & Macro & Seen & Unseen \\
\midrule
Reference
& Teacher Accuracy & 47.82 & 43.78 & 45.19 & 49.74 & 45.55 & 47.22 & 50.53 & 45.90 & 48.18 \\
\midrule
\multirow{4}{*}{Supervised}
& CE & 42.40 & 39.44 & 38.00 & 45.49 & 43.12 & 40.66 & 50.09 & 48.21 & 45.58 \\
& Label smoothing & 41.12 & 37.91 & 36.54 & 44.81 & 42.31 & 40.06 & 49.84 & 47.93 & 45.71 \\
& Mixup & 43.13 & 39.95 & 39.21 & 45.92 & 43.16 & 41.77 & 50.83 & 49.11 & 45.78 \\
& Label smoothing + Mixup & 42.61 & 39.47 & 38.64 & 45.68 & 43.28 & 41.47 & 50.04 & 48.14 & 44.72 \\
\midrule
\multirow{4}{*}{Logit KD}
& Vanilla KD & 45.70 & 42.59 & 40.29 & 49.36 & 47.27 & 45.21 & 53.48 & 51.64 & 49.19 \\
& CTKD & 45.02 & 42.57 & 41.52 & 49.13 & 47.45 & 44.93 & 53.90 & 52.39 & 49.36 \\
& DKD & \underline{46.20} & 42.60 & \textbf{42.66} & \underline{50.16} & \textbf{48.13} & \textbf{47.56} & 54.44 & 52.30 & \textbf{51.00} \\
& BTKD & 44.51 & 41.93 & 42.47 & 49.23 & 47.00 & 45.23 & 53.81 & 52.35 & 49.65 \\
\midrule
\multirow{2}{*}{Structured KD}
& MPD & 45.55 & \underline{42.81} & 41.14 & 49.89 & 47.96 & 45.66 & \underline{54.58} & \underline{52.82} & 50.32 \\
& \textbf{BA-MPD} & \textbf{46.33} & \textbf{43.52} & \underline{42.52} & \textbf{50.76} & \underline{47.99} & \underline{45.87} & \textbf{54.73} & \textbf{53.19} & \underline{50.69} \\
\midrule
& $\Delta$ & +0.63 & +0.93 & +2.23 & +1.40 & +0.72 & +0.66 & +1.25 & +1.55 & +1.50 \\
\bottomrule
\end{tabular}
\end{table*}

\begin{table*}[t]
\centering
\caption{Main results on BEANS-CBI under different low-resource budgets. Reference Teacher denotes the split-specific CNN14 teacher evaluated on the official test split and kept fixed for all KD methods within each budget. Student results are averaged over three random seeds and reported as macro accuracy, macro-F1, and micro accuracy (\%). $\Delta$ denotes the absolute improvement of BA-MPD over vanilla KD.}
\label{tab:beans_main_results}
\footnotesize
\setlength{\tabcolsep}{3.2pt}
\renewcommand{\arraystretch}{1.08}
\begin{tabular}{ccccccccccc}
\toprule
\multirow{3}{*}{Manner} &
\multirow{3}{*}{Method} &
\multicolumn{9}{c}{BEANS-CBI: CNN14 teacher $\rightarrow$ CNN14 student} \\
\cmidrule(lr){3-11}
& & \multicolumn{3}{c}{10\%} & \multicolumn{3}{c}{25\%} & \multicolumn{3}{c}{50\%} \\
\cmidrule(lr){3-5} \cmidrule(lr){6-8} \cmidrule(lr){9-11}
& & Macro & F1 & Micro & Macro & F1 & Micro & Macro & F1 & Micro \\
\midrule
Reference
& Teacher Accuracy
& 6.80 & 6.00 & 7.32
& 19.33 & 17.47 & 21.33
& 37.80 & 35.30 & 40.77 \\
\midrule
\multirow{4}{*}{Supervised}
& CE
& 6.66 & 5.88 & 7.29
& 19.80 & 18.09 & 21.73
& 38.71 & 36.18 & 41.40 \\
& Label smoothing
& 6.44 & 5.65 & 7.15
& 18.72 & 16.75 & 20.52
& 39.28 & 36.87 & 42.04 \\
& Focal loss
& 6.20 & 5.51 & 6.89
& 19.11 & 17.37 & 20.64
& 36.32 & 34.39 & 39.49 \\
& Label smoothing + Focal loss
& 6.34 & 5.37 & 6.91
& 20.41 & 18.47 & 22.21
& 39.17 & 36.74 & 41.98 \\
\midrule
\multirow{4}{*}{Logit KD}
& Vanilla KD
& 7.32 & 6.60 & 7.94
& 22.37 & 20.02 & 24.27
& 40.07 & 37.69 & 43.26 \\
& CTKD
& 7.06 & 6.38 & 8.02
& 22.23 & 20.60 & 24.10
& 41.48 & 39.23 & 44.31 \\
& DKD
& 6.89 & 6.31 & 7.58
& 20.64 & 20.39 & 22.46
& 39.64 & 37.52 & 42.44 \\
& BTKD
& 7.28 & \textbf{6.67} & 7.85
& 22.50 & 20.62 & 24.46
& 41.06 & 38.85 & 44.13 \\
\midrule
\multirow{2}{*}{Structured KD}
& MPD
& \textbf{7.45} & 6.62 & \underline{8.07}
& \textbf{23.08} & \textbf{21.25} & \textbf{25.17}
& \underline{41.79} & \underline{39.58} & \underline{44.70} \\
& \textbf{BA-MPD}
& \underline{7.39} & \underline{6.64} & \textbf{8.12}
& \underline{22.57} & \underline{20.78} & \underline{24.49}
& \textbf{41.91} & \textbf{39.64} & \textbf{44.88} \\
\midrule
& $\Delta$
& +0.07 & +0.04 & +0.18
& +0.20 & +0.76 & +0.22
& +1.84 & +1.95 & +1.62 \\
\bottomrule
\end{tabular}
\end{table*}

\noindent
\textbf{Comparison with supervised regularization.} Tables~\ref{tab:tau_main_results} and~\ref{tab:beans_main_results} compare BA-MPD with dataset-specific supervised regularization baselines and logit-based KD methods. The supervised baselines provide useful but inconsistent gains under low-resource training. On TAU, mixup improves macro accuracy over CE under all budgets, whereas label smoothing is less stable. On BEANS-CBI, focal loss alone underperforms CE across budgets, and label smoothing is beneficial mainly at the 50\% budget. In contrast, BA-MPD consistently outperforms the strongest supervised baseline on the primary metric. These results indicate that structured use of teacher predictions provides gains beyond those obtained from supervised regularization alone.

\noindent
\textbf{Comparison with vanilla KD.} Vanilla KD improves the primary metric over CE under every budget on both benchmarks, showing that teachers trained with limited labels still provide
useful supervision. BA-MPD further improves over vanilla KD across all evaluated budgets, with gains of 0.63, 1.40, and 1.25 percentage points on TAU Urban Acoustic Scenes and 0.07, 0.20, and 1.84 points on BEANS-CBI. These consistent improvements indicate that the proposed boundary correction and separated matching of support relations, support--complement mass, and complement
relations provide a more effective use of teacher predictions than uniform full-distribution matching in the evaluated settings. The contributions of these components are examined separately in the ablation analysis.

\noindent
\textbf{Comparison with logit-based KD methods.} BA-MPD shows strong but metric-dependent performance against the representative logit-based KD methods. On TAU Urban Acoustic Scenes, it achieves the highest macro-averaged accuracy under all three label budgets and the highest
seen-device accuracy at the 5\% and 25\% budgets. DKD performs better on unseen-device accuracy, although BA-MPD still improves over vanilla KD by 2.23, 0.66, and 1.50 percentage points under the three budgets. On BEANS-CBI, BA-MPD outperforms vanilla KD across all reported budgets and metrics. MPD is slightly stronger on several macro-level metrics at the 10\% and 25\% budgets, whereas BA-MPD achieves the best macro accuracy, macro-F1, and micro accuracy at the
50\% budget. Overall, BA-MPD provides consistent improvements over vanilla KD and remains competitive with strong logit-based alternatives, while its relative advantage over other structured objectives varies across tasks, budgets, and evaluation metrics.

\noindent
\textbf{Performance relative to the imperfect teachers.} The fixed teacher results clarify the setting considered in this work. The teacher is not treated as a fully reliable oracle or a performance upper bound, but as a source of informative yet imperfect soft supervision. On TAU, BA-MPD surpasses the teacher macro accuracy at the 10\% and 25\% budgets, reaching 50.76 versus 49.74 and 54.73 versus 50.53, respectively. On BEANS-CBI, BA-MPD also exceeds the teacher macro accuracy across all budgets, improving from 6.80 to 7.39 at 10\%, from 19.33 to 22.57 at 25\%, and from 37.80 to 41.91 at 50\%. These results support the view that the teacher distribution is useful as structured regularization, but should not be treated as a fully reliable target for direct imitation.

\subsection{Ablation and Sensitivity Analysis}

This section examines the design of BA-MPD at both the module and loss-term levels. We first compare vanilla KD, MPD, and BA-MPD to examine the contributions of mass-partitioned distillation and boundary anchoring. We then evaluate the sensitivity to the support size \(k\) and remove individual MPD loss terms to study their respective effects.

\textbf{Component ablation.} Tables~\ref{tab:tau_main_results} and~\ref{tab:beans_main_results} provide a controlled progression from vanilla KD to MPD and BA-MPD. Replacing full-distribution matching with MPD improves the primary metric in five of the six benchmark--budget settings, with the largest gains of 1.10 points on TAU Urban Acoustic Scenes and 1.72 points on BEANS-CBI, both at the highest evaluated budget. Adding BAC to MPD further improves TAU performance by 0.78, 0.87, and 0.15 points under the 5\%, 10\%, and 25\% budgets, respectively. Its effect on BEANS-CBI is more budget-dependent, changing macro accuracy by \(-0.06\), \(-0.51\), and \(+0.12\) points under the 10\%, 25\%, and 50\% budgets. These results show that mass-partitioned distillation provides the more consistent improvement across the two benchmarks, whereas the additional benefit of boundary anchoring depends on the task and label budget.

\textbf{Sensitivity to \(k\).} Table~\ref{tab:k_sensitivity} examines the sensitivity of BA-MPD to the top-\(k\) support size. The default setting used in the main experiments remains strong across both benchmarks: it achieves the best result in five out of six benchmark-budget settings and is within 0.05 points of the best result in the remaining case.

On TAU Urban Acoustic Scenes, the default \(k=2\) consistently performs best across all budgets. Compared with \(k=1\) and \(k=3\), it improves macro accuracy by up to 1.95 and 0.98 points, respectively. This suggests that a compact local support is sufficient for the ten-class acoustic scene task, whereas enlarging the support may introduce less reliable low-probability classes.

On BEANS-CBI, the default \(k=5\) is also stable despite the much larger 264-class label space. It gives the best macro accuracy under the 25\% and 50\% budgets and is only 0.05 points below the best result under the 10\% budget. Increasing \(k\) to 10 or 20 does not provide consistent gains, indicating that broader top-\(k\) supports do not necessarily improve long-tailed bioacoustic recognition. This behavior is consistent with the design of BA-MPD: the anchored top-\(k\) support captures reliable local teacher relations, while complement-region information outside the support is handled through mass-partitioned and complement-region distillation. Overall, BA-MPD is not overly sensitive to the exact choice of \(k\), and the default settings provide a robust balance between support coverage and teacher reliability.

\begin{table}[t]
\centering
\caption{
Top-\(k\) sensitivity of BA-MPD. Results are reported using the primary metric of each benchmark: macro-average scene classification accuracy for TAU Urban Acoustic Scenes and macro accuracy for BEANS-CBI (\%). For TAU, the three settings correspond to \(k=1/2/3\), with default \(k=2\). For BEANS-CBI, they correspond to \(k=5/10/20\), with default \(k=5\). Best results are highlighted in bold, and default settings are underlined.
}
\label{tab:k_sensitivity}
\footnotesize
\setlength{\tabcolsep}{6pt}
\renewcommand{\arraystretch}{1.08}
\begin{tabular}{ccccc}
\toprule
Benchmark & Budget & Setting 1 & Setting 2 & Setting 3 \\
\midrule
\multirow{3}{*}{TAU}
& 5\%  & 45.43 & \underline{\textbf{46.33}} & 45.73 \\
& 10\% & 48.81 & \underline{\textbf{50.76}} & 49.78 \\
& 25\% & 54.05 & \underline{\textbf{54.73}} & 54.45 \\
\midrule
\multirow{3}{*}{BEANS-CBI}
& 10\% & \underline{7.39} & \textbf{7.44} & 7.32 \\
& 25\% & \underline{\textbf{22.57}} & 22.43 & 22.37 \\
& 50\% & \underline{\textbf{41.91}} & 41.83 & 41.67 \\
\bottomrule
\end{tabular}
\end{table}

\textbf{Loss-term ablation.} Table~\ref{tab:loss_term_ablation} investigates the contribution of each loss component in BA-MPD under the default top-\(k\) setting. Overall, the full objective provides strong performance across both benchmarks, showing that mass allocation, top-region relation matching, and complement-region relation matching provide complementary forms of teacher supervision.

\begin{table}[t]
\centering
\caption{
Loss-term ablation of BA-MPD under the default top-\(k\) setting. All variants use the same boundary-anchored support and training protocol, differing only in the distillation terms included. Results are reported using the primary metric of each benchmark: macro-average scene accuracy for TAU Urban Acoustic Scenes and macro accuracy for BEANS-CBI (\%).
}
\label{tab:loss_term_ablation}
\footnotesize
\setlength{\tabcolsep}{2.5pt}
\renewcommand{\arraystretch}{1.05}
\begin{tabular*}{\columnwidth}{@{\extracolsep{\fill}}lccc ccc ccc@{}}
\toprule
\multirow{2}{*}{Variant}
& \multicolumn{3}{c}{Loss Terms}
& \multicolumn{3}{c}{TAU}
& \multicolumn{3}{c}{BEANS-CBI} \\
\cmidrule(lr){2-4}
\cmidrule(lr){5-7}
\cmidrule(lr){8-10}
& \(\mathcal{L}_{\mathrm{mass}}\)
& \(\mathcal{L}_{\mathrm{top}}\)
& \(\mathcal{L}_{\mathrm{comp}}\)
& 5\% & 10\% & 25\%
& 10\% & 25\% & 50\% \\
\midrule
Full BA-MPD
& \(\checkmark\) & \(\checkmark\) & \(\checkmark\)
& \textbf{46.33} & \textbf{50.76} & \textbf{54.73}
& 7.39 & 22.57 & \textbf{41.91} \\
w/o \(\mathcal{L}_{\mathrm{mass}}\)
& -- & \(\checkmark\) & \(\checkmark\)
& 45.53 & 48.97 & 53.70
& 7.45 & 21.84 & 40.86 \\
w/o \(\mathcal{L}_{\mathrm{top}}\)
& \(\checkmark\) & -- & \(\checkmark\)
& 45.08 & 48.39 & 53.46
& \textbf{7.50} & \textbf{23.51} & 41.83 \\
w/o \(\mathcal{L}_{\mathrm{comp}}\)
& \(\checkmark\) & \(\checkmark\) & --
& 44.89 & 48.96 & 54.18
& 6.62 & 20.28 & 38.78 \\
\bottomrule
\end{tabular*}
\end{table}

On TAU Urban Acoustic Scenes, removing any individual term degrades performance across all budgets. This indicates that the three terms contribute complementarily to acoustic scene generalization. The effect is especially clear for \(\mathcal{L}_{\mathrm{top}}\) and \(\mathcal{L}_{\mathrm{comp}}\): removing \(\mathcal{L}_{\mathrm{top}}\) reduces macro accuracy by 1.25, 2.37, and 1.27 points under the 5\%, 10\%, and 25\% budgets, respectively, while removing \(\mathcal{L}_{\mathrm{comp}}\) reduces it by 1.44, 1.80, and
0.55 points. These results suggest that both the anchored top-\(k\) relations and the complement-region information are useful for scene recognition under limited labels.

On BEANS-CBI, the effect of different components is more nuanced due to the larger label space and stronger long-tailed distribution. Removing \(\mathcal{L}_{\mathrm{comp}}\) leads to the most consistent degradation, suggesting that complement-region information is particularly important for long-tailed bioacoustic recognition. Removing \(\mathcal{L}_{\mathrm{mass}}\) also generally weakens performance, indicating that coarse top/complement mass allocation provides useful regularization. In contrast, removing
\(\mathcal{L}_{\mathrm{top}}\) does not always degrade performance, which suggests that normalized top-\(k\) relation matching can be more sensitive when the teacher top-\(k\) support is noisy over a large class space. Nevertheless, the full BA-MPD objective remains competitive across budgets and achieves the best result at the 50\% budget, where the teacher distribution is more reliable.

Overall, these results support the proposed decomposition of teacher supervision: \(\mathcal{L}_{\mathrm{mass}}\) aligns coarse probability allocation, \(\mathcal{L}_{\mathrm{top}}\) captures fine-grained relations within the anchored support, and \(\mathcal{L}_{\mathrm{comp}}\) preserves lower-confidence complement-region information outside the anchored support. The ablation results further indicate that these components provide complementary signals, while the usefulness of normalized top-\(k\) relation matching is more task-dependent in the long-tailed BEANS-CBI setting.

\subsection{Cross-budget Transfer Analysis}

In this section, we study cross-budget distillation under the constraint $r_t \leq r_s$, where the teacher is trained with no more task-specific labeled data than the student. This setting tests whether imperfect low-resource teachers can still provide useful soft supervision, and whether BA-MPD can exploit such signals more reliably than vanilla KD.

\begin{table*}[t]
\centering
\caption{
Cross-budget transfer on TAU Urban Acoustic Scenes with $r_t \leq r_s$. \(r_t\) and \(r_s\) denote the teacher and student label budgets, respectively, and only pairs satisfying \(r_t \le r_s\) are evaluated. Results are reported as macro-average scene classification accuracy (\%). \(\Delta_{\mathrm{CE}}\) denotes the improvement of BA-MPD over the CE student with the same student budget, and \(\Delta_{\mathrm{VKD}}\) denotes the improvement over vanilla KD under the same teacher--student budget pair.
}
\label{tab:tau_cross_budget}
\footnotesize
\setlength{\tabcolsep}{4.2pt}
\renewcommand{\arraystretch}{1.08}
\begin{tabular}{ccccccccc}
\toprule
Teacher Budget \(r_t\)
& Student Budget \(r_s\)
& Teacher Acc.
& CE Student
& Vanilla KD
& MPD
& BA-MPD
& \(\Delta_{\mathrm{CE}}\)
& \(\Delta_{\mathrm{VKD}}\) \\
\midrule
\multirow{3}{*}{5\%}
& 5\%  & \multirow{3}{*}{47.82} & 42.40 & 45.70 & 45.55 & \textbf{46.33} & +3.93 & +0.63 \\
& 10\% &                       & 45.49 & 48.68 & 48.96  & \textbf{49.61} & +4.12 & +0.93 \\
& 25\% &                       & 50.09 & 53.04 & 53.42  & \textbf{53.80} & +3.71 & +0.76 \\
\midrule
\multirow{2}{*}{10\%}
& 10\% & \multirow{2}{*}{49.74} & 45.49 & 49.36 & 49.89 & \textbf{50.76} & +5.27 & +1.40 \\
& 25\% &                        & 50.09 & 54.44 & 54.58 & \textbf{54.72} & +4.63 & +0.28 \\
\midrule
25\%
& 25\% & 50.53 & 50.09 & 53.48 & 54.58 & \textbf{54.73} & +4.64 & +1.25 \\
\bottomrule
\end{tabular}
\end{table*}

\textbf{Cross-budget transfer under controlled label budgets.}
Table~\ref{tab:tau_cross_budget} reports the corresponding cross-budget results on TAU Urban Acoustic Scenes. Across all teacher--student budget pairs, BA-MPD improves over the CE student with the same student budget, with gains ranging from \(+3.71\) to \(+5.27\) points. It also consistently improves over vanilla KD, with \(\Delta_{\mathrm{VKD}}\) ranging from \(+0.28\) to \(+1.40\) points. These results indicate that the benefit of BA-MPD is not limited to the same-budget teacher--student pairs; structurally distilled soft targets remain useful even when the teacher is trained with fewer labels than the student.

\textbf{Lower-budget teachers can still provide useful knowledge.} A key observation is that a lower-budget teacher can improve a higher-budget student when their predictions are distilled structurally. For example, the 5\%-budget teacher achieves only 47.82\% macro accuracy, which is below the 25\%-budget CE student's 50.09\%. Nevertheless, \(T_5\rightarrow S_{25}\) with BA-MPD achieves 53.80, improving over the CE student by \(+3.71\) points and over vanilla KD by \(+0.76\) points. Similarly, \(T_{10}\rightarrow S_{25}\) reaches 54.72, nearly matching the same-budget \(T_{25}\rightarrow S_{25}\) result of 54.73. This suggests that teacher accuracy alone does not determine the usefulness of distillation: lower-budget teachers may still encode transferable class relations that can be exploited when unreliable probability components are handled structurally.

\textbf{Effect of the teacher label budget.} Increasing the teacher budget generally improves cross-budget transfer, but the effect is not linear. With the student fixed at 10\%, increasing the teacher budget from 5\% to 10\% improves BA-MPD from 49.61 to 50.76. With the student fixed at 25\%, increasing the teacher budget from 5\% to 10\% improves BA-MPD from 53.80 to 54.72, but further increasing the teacher budget to 25\% yields only a marginal change to 54.73. This diminishing-return pattern supports the central premise of BA-MPD: once the teacher provides sufficiently informative soft relations, the main challenge is not simply obtaining a stronger teacher, but selectively using the teacher distribution without uniformly imitating its imperfect predictions.

\subsection{Teacher Reliability Analysis}

In this section, we examine and visualize teacher support reliability and its connection to the behavior of structured distillation.

\begin{figure}[t]
\centering
\includegraphics[width=\columnwidth]{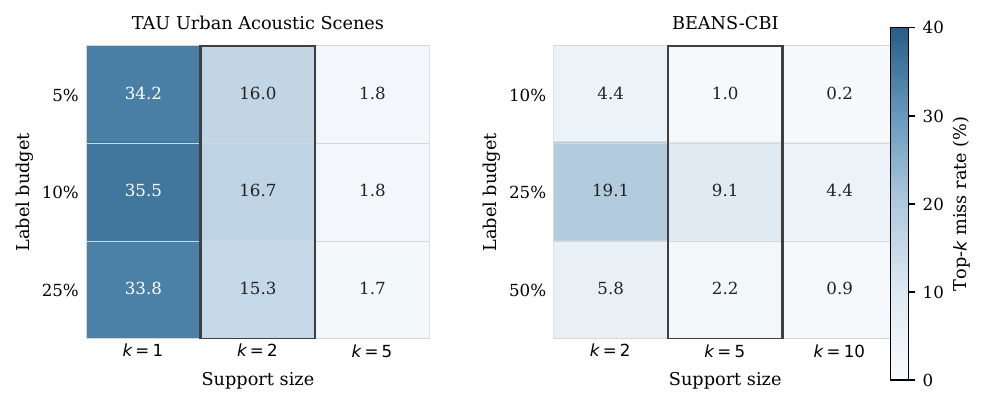}
\caption{
Teacher top-\(k\) support miss rates across support sizes and label budgets.
Each cell reports the percentage of samples whose ground-truth class is excluded from the teacher's top-\(k\) support. Lower values indicate fewer label-support mismatches. Outlined columns mark the default \(k\) used by BA-MPD.
}
\label{fig:topk_support_risk_heatmap}
\end{figure}

\noindent\textbf{Teacher top-\(k\) support risk.} Figure~\ref{fig:topk_support_risk_heatmap} reports the teacher top-\(k\) support miss rate under different support sizes and label budgets. A miss occurs when the ground-truth label is outside the teacher's top-\(k\) support. On TAU Urban Acoustic Scenes, the miss rate is high at \(k=1\), remaining around 33.8--35.5\% across budgets. Increasing the support to the default \(k=2\) substantially reduces the miss rate, but it still remains non-negligible, ranging from 15.3\% to 16.7\%. This confirms that even a compact ten-class acoustic scene task can suffer from support-level teacher errors under low-resource training. Further increasing \(k\) to 5 reduces the miss rate to about 1.7--1.8\%, but such a larger support also introduces more low-probability classes into the distillation region. This motivates boundary anchoring as a conservative alternative to simply enlarging the support.

On BEANS-CBI, the support-risk pattern is more budget-dependent. At the default \(k=5\), the miss rate is low under the 10\% and 50\% budgets, with values of 1.0\% and 2.2\%, respectively, but rises to 9.1\% under the 25\% budget. This indicates that teacher support reliability is not determined solely by the nominal label budget, especially in the 264-class long-tailed bioacoustic setting. Overall, the heatmap supports the motivation of BA-MPD: teacher top-\(k\) support mismatch is a real failure mode, but the severity of this failure mode varies across datasets, budgets, and support sizes.

\begin{figure}[t]
\centering
\includegraphics[scale=0.70]{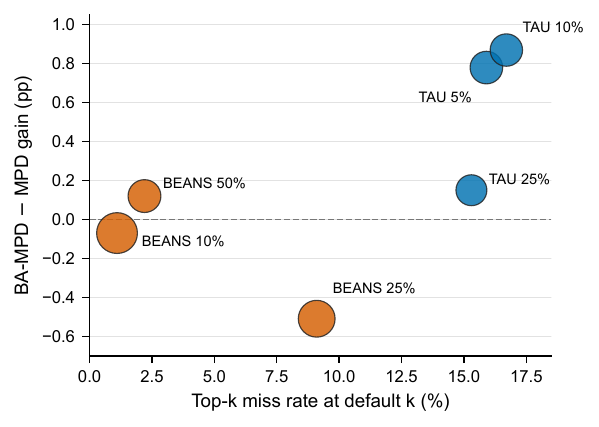}
\caption{
Incremental effect of boundary anchoring across different teacher-support
conditions. The x-axis shows the teacher top-\(k\) support miss rate at the
default \(k\), and the y-axis reports the primary-metric difference between
BA-MPD and MPD. The dashed line marks no performance change, with points above
and below the line indicating positive and negative effects of BAC,
respectively. Bubble area is proportional to the mean teacher probability mass
assigned outside the default top-\(k\) support.
}
\label{fig:anchoring_gain_support_risk}
\end{figure}

\noindent\textbf{Boundary-anchoring gain versus support risk.} Figure~\ref{fig:anchoring_gain_support_risk} relates the default-\(k\) support
miss rate to the incremental effect of BAC over MPD. On TAU Urban Acoustic
Scenes, adding BAC to MPD produces positive average differences of 0.78, 0.87,
and 0.15 percentage points under the 5\%, 10\%, and 25\% label budgets,
respectively. The corresponding support miss rates range from approximately
15\% to 17\%. This pattern is consistent with the intended role of BAC, which
targets training samples whose ground-truth class is excluded from the
teacher's top-\(k\) support.

The effect on BEANS-CBI is more budget-dependent. The average differences
between BA-MPD and MPD are approximately \(-0.07\), \(-0.51\), and \(+0.12\)
percentage points under the 10\%, 25\%, and 50\% budgets, respectively. In
particular, the largest support miss rate does not correspond to a larger BAC
improvement, showing that miss frequency alone is insufficient to determine the
benefit of boundary anchoring. The bubble sizes also show that a non-negligible
fraction of teacher probability mass can remain outside the default top-\(k\)
support. These observations distinguish the roles of the two components: MPD
provides structured transfer of support relations, support--complement mass,
and complement information, whereas BAC provides an additional targeted
correction whose effect depends on the teacher-error structure of the task.

\section{Conclusion}
This paper studied knowledge distillation for low-resource acoustic representation learning with informative but imperfect teachers. We proposed Boundary-Anchored Mass-Partitioned Distillation (BA-MPD), which treats teacher predictions as structured supervision rather than as uniformly reliable targets. Specifically, BA-MPD mitigates label-support mismatch through boundary-anchored correction and decomposes teacher supervision into mass allocation, anchored-support relations, and complement-region information. Experiments on acoustic scene and bioacoustic recognition benchmarks show that BA-MPD consistently improves over CE and vanilla KD, while remaining competitive with representative logit-based KD methods under different low-resource budgets. The experimental results further demonstrate that imperfect teachers can still provide useful dark knowledge when their predictions are structurally distilled, highlighting the importance of selective teacher-distribution transfer in low-resource acoustic learning.

\section{Generative AI Use Disclosure}
The authors acknowledge the use of GPT 5.5 tools during manuscript preparation for language editing and readability improvement. The core methodology, algorithmic implementation, and evaluations were developed, executed, and verified by the authors. The authors reviewed and revised all AI-assisted content and take full responsibility for the final submission.


\bibliographystyle{IEEEtran}
\bibliography{taslp}

@String(CVPR= {IEEE Conf. Comput. Vis. Pattern Recog.})

@String(ICASSP=	{ICASSP})

@String(AAAI = {AAAI})

@String(CVPR  = {CVPR})

@article{stanton2021does,
  title={Does knowledge distillation really work?},
  author={Stanton, Samuel and Izmailov, Pavel and Kirichenko, Polina and others},
  journal={Advances in neural information processing systems},
  volume={34},
  pages={6906--6919},
  year={2021}
}

@article{syed2024data,
  author  = {Ali Raza Syed and Enis Berk {\c{C}}oban and Dara Pir and Michael I. Mandel},
  title   = {Data-Centric Methods for Environmental Sound Classification With Limited Labels},
  journal = {IEEE/ACM Transactions on Audio, Speech, and Language Processing},
  volume  = {32},
  pages   = {4288--4297},
  year    = {2024},
  doi     = {10.1109/TASLP.2024.3414332}
}

@article{koutini2021receptive,
  author  = {Khaled Koutini and Hamid Eghbal-zadeh and Gerhard Widmer},
  title   = {Receptive Field Regularization Techniques for Audio Classification and Tagging With Deep Convolutional Neural Networks},
  journal = {IEEE/ACM Transactions on Audio, Speech, and Language Processing},
  volume  = {29},
  pages   = {1987--2000},
  year    = {2021},
  doi     = {10.1109/TASLP.2021.3082307}
}

@inproceedings{heittola2020dcase,
  author    = {Toni Heittola and Annamaria Mesaros and Tuomas Virtanen},
  title     = {Acoustic Scene Classification in {DCASE} 2020 Challenge: Generalization Across Devices and Low Complexity Solutions},
  booktitle = {Proceedings of the Detection and Classification of Acoustic Scenes and Events 2020 Workshop ({DCASE} 2020)},
  pages     = {56--60},
  year      = {2020},
  url       = {https://arxiv.org/abs/2005.14623}
}

@inproceedings{martin2021dcase,
  author    = {Irene Mart{\'i}n-Morat{\'o} and Toni Heittola and Annamaria Mesaros and Tuomas Virtanen},
  title     = {Low-Complexity Acoustic Scene Classification for Multi-Device Audio: Analysis of {DCASE} 2021 Challenge Systems},
  booktitle = {Proceedings of the Detection and Classification of Acoustic Scenes and Events 2021 Workshop ({DCASE} 2021)},
  pages     = {85--89},
  year      = {2021},
  doi       = {10.5281/zenodo.5770113}
}

@inproceedings{kim2022rfn,
  author    = {Byeonggeun Kim and Seunghan Yang and Jangho Kim and Hyunsin Park and Juntae Lee and Simyung Chang},
  title     = {Domain Generalization with Relaxed Instance Frequency-wise Normalization for Multi-device Acoustic Scene Classification},
  booktitle = {Interspeech 2022},
  pages     = {2393--2397},
  year      = {2022},
  doi       = {10.21437/Interspeech.2022-61}
}

@inproceedings{hagiwara2023beans,
  author    = {Masato Hagiwara and Benjamin Hoffman and Jen-Yu Liu and Maddie Cusimano and Felix Effenberger and Katie Zacarian},
  title     = {{BEANS}: The Benchmark of Animal Sounds},
  booktitle = {Proceedings of the IEEE International Conference on Acoustics, Speech and Signal Processing ({ICASSP})},
  year      = {2023},
  doi       = {10.1109/ICASSP49357.2023.10095776}
}

@inproceedings{hinton2015distilling,
  author    = {Geoffrey Hinton and Oriol Vinyals and Jeff Dean},
  title     = {Distilling the Knowledge in a Neural Network},
  booktitle = {NeurIPS},
  year      = {2015},
  url       = {https://arxiv.org/abs/1503.02531}
}

@article{gou2021knowledge,
  author  = {Jianping Gou and Baosheng Yu and Stephen J. Maybank and Dacheng Tao},
  title   = {Knowledge Distillation: A Survey},
  journal = {International Journal of Computer Vision},
  volume  = {129},
  number  = {6},
  pages   = {1789--1819},
  year    = {2021},
  doi     = {10.1007/s11263-021-01453-z}
}

@inproceedings{zhao2022decoupled,
  author    = {Borui Zhao and Quan Cui and Renjie Song and Yiyu Qiu and Jiajun Liang},
  title     = {Decoupled Knowledge Distillation},
  booktitle = {Proceedings of the IEEE/CVF Conference on Computer Vision and Pattern Recognition},
  pages     = {11943--11952},
  year      = {2022},
  doi       = {10.1109/CVPR52688.2022.01165}
}

@inproceedings{zhou2021rethinking,
  author    = {Helong Zhou and Liangchen Song and Jiajie Chen and Ye Zhou and Guoli Wang and Junsong Yuan and Qian Zhang},
  title     = {Rethinking Soft Labels for Knowledge Distillation: A Bias-Variance Tradeoff Perspective},
  booktitle = {International Conference on Learning Representations},
  year      = {2021},
  url       = {https://openreview.net/forum?id=gIHd-5X324}
}

@inproceedings{lukasik2021teachers,
  author    = {Michal Lukasik and Srinadh Bhojanapalli and Aditya Krishna Menon and Sanjiv Kumar},
  title     = {Teacher's Pet: Understanding and Mitigating Biases in Distillation},
  booktitle = {NeurIPS 2021 Workshop on Distribution Shifts},
  year      = {2021},
  url       = {https://arxiv.org/abs/2106.10494}
}

@article{zhang2023blindly,
  author  = {Rongzhi Zhang and Jiaming Shen and Tianqi Liu and Jialu Liu and Michael Bendersky and Marc Najork and Chao Zhang},
  title   = {Do Not Blindly Imitate the Teacher: Using Perturbed Loss for Knowledge Distillation},
  journal = {arXiv preprint arXiv:2305.05010},
  year    = {2023},
  url     = {https://arxiv.org/abs/2305.05010}
}

@inproceedings{mishra2026correctness,
  author    = {Ishan Mishra and Deepak Mishra and Jinjun Xiong},
  title     = {Correctness-Aware Knowledge Distillation for Enhanced Student Learning},
  booktitle = {Transactions on Machine Learning Research},
  year      = {2026},
  url       = {https://openreview.net/forum?id=XpRXmzd2sF}
}

@misc{schmid2024dcase,
  author       = {Florian Schmid and Paul Primus and Toni Heittola and Annamaria Mesaros and Irene Mart{\'i}n-Morat{\'o} and Khaled Koutini and Gerhard Widmer},
  title        = {Data-Efficient Low-Complexity Acoustic Scene Classification in the {DCASE} 2024 Challenge},
  year         = {2024},
  eprint       = {2405.10018},
  archivePrefix = {arXiv},
  primaryClass = {cs.SD},
  url          = {https://arxiv.org/abs/2405.10018}
}

@article{kong2020panns,
  title={{PANN}s: Large-Scale Pretrained Audio Neural Networks for Audio Pattern Recognition},
  author={Kong, Qiuqiang and Cao, Yin and Iqbal, Turab and Wang, Yuxuan and Wang, Wenwu and Plumbley, Mark D.},
  journal={IEEE/ACM Transactions on Audio, Speech, and Language Processing},
  volume={28},
  pages={2880--2894},
  year={2020}
}

@inproceedings{koutini2021passt,
  title={Efficient Training of Audio Transformers with Patchout},
  author={Koutini, Khaled and Schl{\"u}ter, Jan and Eghbal-zadeh, Hamid and Widmer, Gerhard},
  booktitle={Interspeech},
  pages={2753--2757},
  year={2022}
}

@inproceedings{li2023curriculum,
  title={Curriculum Temperature for Knowledge Distillation},
  author={Li, Zheng and Li, Xiang and Yang, Lingfeng and Zhao, Borui and Song, Renjie and Luo, Lei and Li, Jun and Yang, Jian},
  booktitle={Proceedings of the AAAI Conference on Artificial Intelligence},
  year={2023}
}

@article{zhang2024can,
  title={Can Students Beyond The Teacher? Distilling Knowledge from Teacher's Bias},
  author={Zhang, Jianhua and Gao, Yi and Liu, Ruyu and Cheng, Xu and Zhang, Houxiang and Chen, Shengyong},
  journal={Proceedings of the AAAI Conference on Artificial Intelligence},
  year={2025}
}

@inproceedings{romero2015fitnets,
  title={FitNets: Hints for Thin Deep Nets},
  author={Romero, Adriana and Ballas, Nicolas and Kahou, Samira Ebrahimi and Chassang, Antoine and Gatta, Carlo and Bengio, Yoshua},
  booktitle={International Conference on Learning Representations},
  year={2015}
}

@inproceedings{szegedy2016rethinking,
  title={Rethinking the Inception Architecture for Computer Vision},
  author={Szegedy, Christian and Vanhoucke, Vincent and Ioffe, Sergey and Shlens, Jonathon and Wojna, Zbigniew},
  booktitle={Proceedings of the IEEE Conference on Computer Vision and Pattern Recognition},
  pages={2818--2826},
  year={2016}
}

@inproceedings{zhang2018mixup,
  title={Mixup: Beyond Empirical Risk Minimization},
  author={Zhang, Hongyi and Cisse, Moustapha and Dauphin, Yann N. and Lopez-Paz, David},
  booktitle={International Conference on Learning Representations},
  year={2018}
}

@inproceedings{lin2017focal,
  title={Focal Loss for Dense Object Detection},
  author={Lin, Tsung-Yi and Goyal, Priya and Girshick, Ross and He, Kaiming and Doll{\'a}r, Piotr},
  booktitle={Proceedings of the IEEE International Conference on Computer Vision},
  pages={2980--2988},
  year={2017}
}

@inproceedings{park2019relational,
  title={Relational Knowledge Distillation},
  author={Park, Wonpyo and Kim, Dongju and Lu, Yan and Cho, Minsu},
  booktitle={Proceedings of the IEEE/CVF Conference on Computer Vision and Pattern Recognition},
  pages={3967--3976},
  year={2019}
}

@inproceedings{loshchilov2019decoupled,
  title={Decoupled Weight Decay Regularization},
  author={Loshchilov, Ilya and Hutter, Frank},
  booktitle={International Conference on Learning Representations},
  year={2019}
}

@inproceedings{schmid2023distilling,
  title={Distilling the Knowledge of Transformers and CNNs with CP-Mobile},
  author={Schmid, Florian and Morocutti, Tobias and Masoudian, Shahed and Koutini, Khaled and Widmer, Gerhard},
  booktitle={Proceedings of the Detection and Classification of Acoustic Scenes and Events Workshop (DCASE)},
  pages={161--165},
  year={2023}
}

@inproceedings{baevski2020wav2vec,
  title     = {{wav2vec} 2.0: A Framework for Self-Supervised Learning of Speech Representations},
  author    = {Baevski, Alexei and Zhou, Henry and Mohamed, Abdelrahman and Auli, Michael},
  booktitle = {Advances in Neural Information Processing Systems},
  volume    = {33},
  pages     = {12449--12460},
  year      = {2020}
}

@article{hsu2021hubert,
  title   = {{HuBERT}: Self-Supervised Speech Representation Learning by Masked Prediction of Hidden Units},
  author  = {Hsu, Wei-Ning and Bolte, Benjamin and Tsai, Yao-Hung Hubert and Lakhotia, Kushal and Salakhutdinov, Ruslan and Mohamed, Abdelrahman},
  journal = {IEEE/ACM Transactions on Audio, Speech, and Language Processing},
  volume  = {29},
  pages   = {3451--3460},
  year    = {2021},
  doi     = {10.1109/TASLP.2021.3122291}
}

@inproceedings{chen2023beats,
  title     = {{BEATs}: Audio Pre-Training with Acoustic Tokenizers},
  author    = {Chen, Sanyuan and Wu, Yu and Wang, Chengyi and Liu, Shujie and Tompkins, Daniel and Chen, Zhuo and Che, Wanxiang and Yu, Xiangzhan and Wei, Furu},
  booktitle = {Proceedings of the 40th International Conference on Machine Learning},
  series    = {Proceedings of Machine Learning Research},
  volume    = {202},
  pages     = {5178--5193},
  year      = {2023},
  publisher = {PMLR}
}

@inproceedings{cui2019class,
  title={Class-Balanced Loss Based on Effective Number of Samples},
  author={Cui, Yin and Jia, Menglin and Lin, Tsung-Yi and Song, Yang and Belongie, Serge},
  booktitle={CVPR},
  pages={9268--9277},
  year={2019}
}

@article{zhang2025systematic,
  title={A systematic review on long-tailed learning},
  author={Zhang, Chongsheng and Almpanidis, George and Fan, Gaojuan and Deng, Binquan and Zhang, Yanbo and Liu, Ji and Kamel, Aouaidjia and Soda, Paolo and Gama, Jo{\~a}o},
  journal={IEEE Transactions on Neural Networks and Learning Systems},
  year={2025},
  publisher={IEEE}
}

@inproceedings{huang2025learn,
  title={Learn from balance: Rectifying knowledge transfer for long-tailed scenarios},
  author={Huang, Xinlei and Tang, Jialiang and Zheng, Xubin and Zhou, Jinjia and Yu, Wenxin and Jiang, Ning},
  booktitle={ICASSP},
  pages={1--5},
  year={2025},
  organization={IEEE}
}

@article{chen2022htsat,
  title={HTS-AT: A hierarchical token-semantic audio transformer for sound classification and detection},
  author={Chen, Ke and Du, Xingjian and Zhu, Bilei and Ma, Zejun and Berg-Kirkpatrick, Taylor and Dubnov, Shlomo},
  journal={ICASSP},
  pages={646--650},
  year={2022},
  organization={IEEE}
}

\vfill

\end{document}